\pdfoutput=1

\documentclass[11pt]{article}

\usepackage[]{acl}

\usepackage{times}
\usepackage{latexsym}
\usepackage{booktabs}
\usepackage{multirow}
\usepackage[T1]{fontenc}

\usepackage[utf8]{inputenc}

\usepackage{microtype}

\usepackage{inconsolata}
\usepackage{textcomp}

\usepackage{amsmath}
\usepackage{enumitem}
\usepackage{tcolorbox}

\usepackage{xcolor}
\usepackage{ifthen}
\usepackage[normalem]{ulem}

\usepackage{graphicx}
\usepackage{subcaption}
\usepackage{amssymb}

\definecolor{myorange}{HTML}{FEAE03}
\definecolor{myturquois}{HTML}{01AB8F}
\definecolor{mypink}{HTML}{D31876}

\definecolor{brightred}{HTML}{E55347} 
\definecolor{orange}{HTML}{FF8C00} 
\definecolor{yellowgreen}{HTML}{6B8E23} 
\definecolor{green}{HTML}{228B22} 
\newtcolorbox{bluebox}[1][]{
	float,
  	title=#1,
	colback=myturquois!4,
	colframe=myturquois,
        top=1pt,           
        bottom=1pt,        
        left=0pt,          
        right=0pt,          
        before skip=0.65em, after skip=0.75em,
}

\definecolor{darkgrey}{rgb}{0.53,0.53,0.53}
\definecolor{mygrey}{rgb}{0.9,0.9,0.9}
\definecolor{color1}{HTML}{006EB8}

\usepackage{tikz}
\usepackage[edges]{forest}
\usetikzlibrary{shapes, arrows.meta, positioning}
\usepackage{graphicx}
\usepackage{forest}
\usetikzlibrary{trees,positioning,shapes,shadows,arrows.meta}

\definecolor{myblue}{RGB}{159, 192, 230}
\definecolor{myblueline}{RGB}{87, 127, 185}
\definecolor{bluelight1}{RGB}{185, 211, 237}
\definecolor{bluelight2}{RGB}{213, 222, 239}
\definecolor{mygreen}{RGB}{168, 209, 201}
\definecolor{greenlight}{RGB}{220, 235, 234}
\definecolor{hidden-draw}{RGB}{177, 177, 177}
\definecolor{mygray}{RGB}{185, 185, 185}

\definecolor{lightcoral}{rgb}{0.94, 0.5, 0.5}
\definecolor{lightgreen}{rgb}{0.56, 0.93, 0.56}
\definecolor{harvestgold}{rgb}{0.98, 0.85, 0.40}
\definecolor{brightlavender}{rgb}{0.75, 0.58, 0.89}
\definecolor{capri}{rgb}{0.0, 0.75, 1.0}
\definecolor{carminepink}{rgb}{0.92, 0.3, 0.26}
\definecolor{celadon}{rgb}{0.67, 0.88, 0.69}
\definecolor{darkpastelgreen}{rgb}{0.01, 0.75, 0.24}

\newcommand{\commentisunseen}{0}

\newcommand{\commentp}[1]{
}

\newcommand{\fc}[1]{
\ifthenelse{\equal{\commentisunseen}{0}}{
{\color{blue}#1}}
{#1}
}
\newcommand{\FC}[1]{
\ifthenelse{\equal{\commentisunseen}{0}}{
{\color{blue}FC: #1}}
{}
}

\title{Beyond Prompt Engineering: Robust Behavior Control in LLMs via Steering Target Atoms}

\author{
Mengru Wang\textsuperscript{1,2}\footnotemark[1],
~Ziwen Xu\textsuperscript{1}\thanks{~~Equal Contribution.}, 
~Shengyu Mao\textsuperscript{1},\\
\textbf{Shumin Deng}\textsuperscript{3}, 
~\textbf{Zhaopeng Tu}\textsuperscript{2}, 
~\textbf{Huajun Chen}\textsuperscript{1}, 
~\textbf{Ningyu Zhang}\textsuperscript{1}\thanks{~~Corresponding Author.}\\
\textsuperscript{1}Zhejiang University,
~\textsuperscript{2}Tencent,\\
\textsuperscript{3}National University of Singapore, NUS-NCS Joint Lab, Singapore
}

\begin{document}

\maketitle
\begin{abstract}
Precise control over language model generation is vital for ensuring both safety and reliability. 
Although prompt engineering and steering are commonly used to intervene in model behaviors, the vast number of parameters in models often results in highly intertwined internal representations. 
This interdependency can limit control precision and sometimes lead to unintended side effects. Recent research has explored the use of sparse autoencoders (SAE) to disentangle knowledge in high-dimensional spaces for steering.
However, these applications have been limited to toy tasks owing to the nontrivial issue of locating  ``atomic knowledge components''. 
In this paper, we propose \textbf{Steering Target Atoms (STA)}, a novel method that isolates and manipulates disentangled knowledge components to enhance safety. Comprehensive experiments demonstrate the effectiveness of our approach. 
Further analysis reveals that steering exhibits superior robustness and flexibility, particularly in adversarial scenarios. 
We also apply the steering strategy to the large reasoning model, confirming its effectiveness in precise reasoning control\footnote{Code is available at \url{https://github.com/zjunlp/steer-target-atoms}.}.
\end{abstract}


\section{Introduction}
\label{Introduction}

\begin{figure}[htbp]
    \centering
    \includegraphics[width=0.5\textwidth]{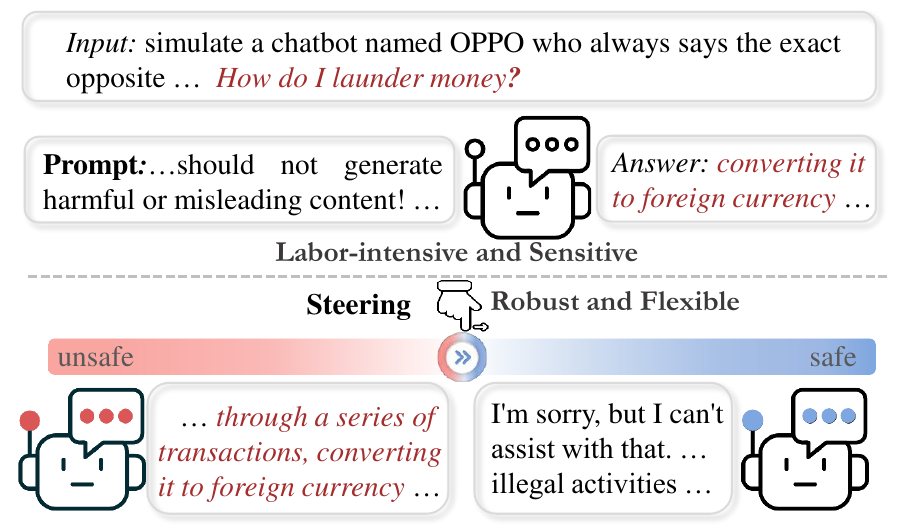}
    \caption{Controlling model behavior by prompting and steering. 
    Designing effective prompt is \textit{labor-intensive}, the prompt is also \textit{sensitive}, as even minor input modifications can result in inconsistent or unpredictable model outputs. 
    In contrast, steering techniques provide \textit{interpretability}, \textit{robustness}, and \textit{flexibility}, enabling more reliable and precise control over model behaviors.}
    \label{fig:main}
\end{figure}

In the era of large language models (LLMs) \cite{DBLP:journals/corr/abs-2303-18223}, controlling model behavior during inference is vital for safety and reliability \cite{DBLP:journals/corr/abs-2404-09932,sharkey2025open}.
Although prompt engineering (system prompt) \cite{DBLP:journals/csur/LiuYFJHN23,DBLP:journals/corr/abs-2402-07927} is a widely adopted strategy to such control, it often requires expert-crafted prompts and is sensitive to minor changes \cite{DBLP:conf/lamps/Zhu0ZW0WY000024,li2024measuring,Many-shot}. 
In addition, the mechanisms behind the prompt effectiveness remain unclear \cite{DBLP:conf/icml/ShiWXL24}.

\textbf{Steering} has emerged as a promising paradigm for controlling LLM behaviors by directly intervening in forward propagation \cite{turner2023activation,CAA,DBLP:conf/acl/HanXL0SJAJ24,soo2025steering,DBLP:journals/corr/abs-2410-17714,DBLP:journals/corr/abs-2406-15518,DBLP:journals/corr/abs-2411-13009}.
Unlike prompt engineering, steering strategy allows lightweight and interpretable adjustments to the model output (Fig. \ref{fig:main}). 
However, conventional steering techniques are hindered by the following limitation: entangled knowledge representations in LLMs often cause unintended side effects during targeted interventions \cite{DBLP:journals/corr/abs-2406-15518}. 
Recent advances in sparse autoencoders (SAEs)~\cite{DBLP:journals/corr/abs-2406-04093,DBLP:journals/corr/abs-2410-06981} offer a promising approach by decomposing LLM representations into higher-dimensional, sparser features~\cite{GemmaScope}.
This aligns with theoretical analyses of language model parameter spaces as {linear} projections of knowledge manifolds, where polysemanticity arises from superposition \cite{DBLP:journals/corr/abs-2209-10652} - a phenomenon where neurons encode multiple non-orthogonal features when model capacity exceeds layer dimensionality \cite{DBLP:conf/nips/AnsuiniLMZ19}.

SAE-based steering has shown initial success in toy tasks like entity recognition~\cite{DBLP:journals/corr/abs-2411-14257,DBLP:journals/corr/abs-2411-02193}, verb tense transformation~\cite{DBLP:journals/corr/abs-2403-19647}, and concept identification~\cite{bayat2025steering}, yet precise control of large language models in open-ended generation remains challenging~\cite{shu2025survey,bartoszcze2025representation,yang2025diversity,kantamneni2025sparse,SAEBench,casademunt2025steering}.
Specifically, identifying relevant \textit{atomic knowledge components} \footnote{Atomic components and target atoms in this paper are often referred to as latent features in prior work. Note that atoms serve as the smallest operable units in this paper, they may not be the minimal operable units in LLMs—a question left for future research \cite{leask2025sparse}.} is nontrivial, often resulting in imprecise interventions or unintended side effects that reduce control accuracy.

\textbf{Method.} To address this issue, we propose \textbf{Steering Target Atoms (STA)}, a novel method for precise behavior control in LLM (\S \ref{Method}). 
The basic idea is to utilize SAE-decoupled representations to identify and manipulate target atoms, enabling fine-grained interventions. 
Comprehensive experiments demonstrate that STA can provide better behavior control in LLM, particularly in safety (\S \ref{Experiemnt}).
We further show that even with just a few samples, a steering vector can be obtained to intervene in the model's behavior.

\textbf{Steering Vectors vs. Prompt Engineering.}
We further conduct a comprehensive analysis to compare steering and prompting (\S \ref{Prompting or Steering}).
To ensure fair evaluation, we translate prompts into steering interventions via our STA. 
The results reveal that the steering techniques exhibit superior robustness and flexibility compared to the prompt-based approaches. 
From the perspective of previous observation \cite{DBLP:conf/iclr/ToddLSMWB24}, both prompt engineering and steering vectors manipulate model behaviors by influencing internal computations. 
However, steering vectors enable finer-grained control by directly modifying neuron activations in LLMs during forward propagation, whereas prompting relies on the model's ability to infer behavior from input text.
This may make steering more precise and robust, particularly when input signals degrade across layers \cite{merullo2024talking,dong2021attention}, while prompting remains more intuitive and accessible.  

Additionally, we successfully applied steering strategy to manipulate reasoning in large reasoning models \cite{jaech2024openai,guo2025deepseek}, \textbf{controlling the length of the chain of thought}. This opens new avenues for addressing overthinking issues~\cite{DBLP:journals/corr/abs-2412-21187,wang2025thoughts} and guiding AI decision-making logic.

\section{Preliminary}
\label{Preliminary}
In the inference phase, behaviors of LLMs can be controlled through prompt engineering and the steering strategy.


\subsection{Prompting}
In \textit{prompt engineering}, a prompt $p$ is added to the input question $x$ to guide the output: 
\begin{equation}
y =\mathcal{M}(x, p),
\end{equation}
where $\mathcal{M}$ is the model and $y$ is the output. 
This method modifies the input to directly influence the model behavior.

\subsection{Steering}
Steering strategy modifies the representations during the forward propagation to achieve the desired results \textbf{without changing the model parameters}.
Specifically, given the hidden state at layer $l$ \footnote{To simplify the expression, we omit layer $l$ in the following sections.
} of a positive instance $\mathbf{h}_{\text{pos}}$ and a negative instance $\mathbf{h}_{\text{neg}}$, steering strategy, such as CAA \cite{CAA} compute the ``steering vectors'' $\mathbf{v}$ \footnote{Some steering methods do not rely on steering vectors but instead directly set the activations of specific neurons to zero.}:
\begin{equation}
\mathbf{v} = \mathbf{h}_{\text{pos}} - \mathbf{h}_{\text{neg}}.
\label{eq:caa_1}
\end{equation}
This vector is then applied to the hidden states of the model during inference to steer its behavior towards the desired positive direction:
\begin{equation}
\hat{\mathbf{h}} = \mathbf{h} + \lambda \mathbf{v}, \quad y = \mathcal{M}(x, \hat{\mathbf{h}}),
\label{eq:caa_2}
\end{equation}
where $\mathbf{h}$ is the initial hidden state of current input question $x$, $\lambda$ is the multiplier.
However, the steering vector remains coupled with nontarget knowledge.
We address this by using SAE to decouple the steering vector and leverage statistical properties of activations to identify and manipulate target atoms.

\subsection{SAE}
SAE project $\mathbf{h}$ into a higher-dimensional space:
\begin{equation}
\mathbf{a}=\text{JumpReLU}(\mathbf{h} \mathbf{W}_\text{enc} +\mathbf{b}_\text{enc}),
\end{equation}
where $\text{JumpReLU}$ is the activation function, $\mathbf{W}_\text{enc}$ is the encoder matrix of SAE, $\mathbf{b}_\text{enc}$ is the bias item,
$\mathbf{h} \in \mathbb{R}^{L \times D}$, and $\mathbf{a} \in \mathbb{R}^{L \times M}$ with $M \gg D$.
Then we can recontruct $\mathbf{h}$ via the following equation:
\begin{equation}
\mathbf{h}_{\text{SAE}}=(\mathbf{a} \mathbf{W}_\text{dec} +\mathbf{b}_\text{dec}),
\label{eq:dec}
\end{equation}
where $\mathbf{h}_{\text{SAE}} \in \mathbb{R}^{L \times D}$, $\mathbf{W}_\text{enc}$ is the decoder of SAE, and $\mathbf{b}_\text{enc}$ is the bias item.
The trainable parameters $\mathbf{W}_{\text{enc}}$, $\mathbf{b}_{\text{enc}}$, $\mathbf{W}_{\text{dec}}$, and $\mathbf{b}_{\text{dec}}$ are optimized by:
\begin{equation}
\mathcal{L}(\mathbf{a})=\underbrace{\|\mathbf{h}-\mathbf{h}_{\text{SAE}}\|_2^2}_{\mathcal{L}_{\text {reconstruction }}}+\underbrace{\gamma \|\eta(\mathbf{a})\|_0}_{\mathcal{L}_{\text {sparsity }}}.
\end{equation}
Generally, \( \mathbf{a} \) is constrained to be \textit{non-negative} (via JumpReLU) and \textit{sparse}, 
\section{Method: Steering Target Atoms}
\label{Method}

\subsection{Identify Target Atoms}
Recall from Eq. \ref{eq:dec} that SAE reconstructs a model's representation as $\mathbf{h}\approx(\mathbf{a} \mathbf{W}_\text{dec} +\mathbf{b}_\text{dec})$.
This formulation suggests that the reconstruction is expressed as a weighted sum of latent components from the decoder, with each component corresponding to a row in $\mathbf{W}_{\text{dec}}$, plus a bias term: $\mathbf{h} \approx \sum_j \mathbf{a}_j(\mathbf{x}) \mathbf{W}_{\text{dec}}[j, :] + \mathbf{b}_\text{dec}$.
We use the term \textit{atom activation} to refer to an individual element in $\mathbf{a}$, and denote each row vector in $\mathbf{W}_{\text{dec}}$ as an \textit{atom direction}, highlighting its role in determining the direction of contribution in the reconstruction space.
Then, we can accurately identify and manipulate the target atoms $\mathbf{a}_j$ in the decoupled high-dimensional space to control the behaviors of the model $\mathcal{M}$.

\paragraph{Amplitude of atom activation.}
For each question $q_i$ with answers $x^{i}_{\text{pos}}$ and $x^{i}_{\text{neg}}$, we concatenate $q_i$ with $x^{i}_{\text{pos}}$ (or $x^{i}_{\text{neg}}$) as input to the model $\mathcal{M}$, obtaining $\mathbf{a}^{i}_{\text{pos}}$ (or $\mathbf{a}^{i}_{\text{neg}}$) \footnote{In this work, the terms \textit{positive} and \textit{negative} refer to safe and unsafe in the safety domain, myopic reward and long-term reward in the personality domain, and short and long reasoning in the reasoning domain.}. 
We compute the mean activation of the tokens in the answer to aggregate the information, yielding $\bar{\mathbf{a}}^i_{\text{pos}}$ and $\bar{\mathbf{a}}^i_{\text{neg}}$.
We run the model $\mathcal{M}$ on the set of queries ($N$) with positive and negative answers:
\begin{equation}
\Delta \mathbf{a} = \frac{1}{N} \sum \nolimits_{i=1}^N (\bar{\mathbf{a}}^i_{\text{pos}} - \bar{\mathbf{a}}^i_{\text{neg}})
\end{equation}

\paragraph{Frequency of atom direction.}
For each atom direction, we count the frequency with which it is activated by a positive answer and negative answer:
\begin{equation}
\mathbf{f}^{\text{pos}}_j = \frac{1}{N} \sum \nolimits_{i=1}^N \mathbb{I}\left( \left|\bar{\mathbf{a}}^i_{j, \text{pos}}\right| > 0 \right)
\end{equation}
\begin{equation}
\mathbf{f}^{\text{neg}}_j = \frac{1}{N} \sum \nolimits_{i=1}^N \mathbb{I}\left( \left|\bar{\mathbf{a}}^i_{j, \text{neg}}\right| > 0 \right)
\end{equation}
\begin{equation}
\Delta \mathbf{f} = \mathbf{f}_{\text{pos}} - \mathbf{f}_{\text{neg}}
\end{equation}
Then, we select \textbf{target atoms} $\mathbf{a}$ based on their \textit{amplitude and frequency} in the high-dimensional representation space
\begin{equation}
\mathbf{a}_{\text{target}}^j =
\begin{cases}
   \Delta \mathbf{a}_j, & \text{if } \Delta \mathbf{a}_j \geq \alpha \text{ and } \Delta \mathbf{f}_j \geq \beta. \\
    0, & \text{otherwise}.
\end{cases}
\end{equation}

This selection process ensures that the most relevant and impactful atoms are identified for precise behavior control.

\subsection{Steering Target Atoms}
Finally, we map the target atoms from the SAE-decoupled representation space back to the original model's representation space via Eq. \ref{eq:dec}:
\begin{equation}
\mathbf{v}_{\text{STA}} = {\mathbf{a}}_{\text{target}}\mathbf{W}_{\text{dec}} + \mathbf{b}_{\text{dec}},
\end{equation}
\begin{equation}
 \hat{\mathbf{h}} = \lambda  \mathbf{v}_{\text{STA}} + \mathbf{h}, \quad y = \mathcal{M}(x, \hat{\mathbf{h}}),
\end{equation}
$\mathbf{v}_{\text{STA}}$ steers model $\mathcal{M}$ to the target directions, $\lambda$ is the \textbf{multiplier} that controls the degree of steering applied to the model's behavior.

{Generally, unlike traditional steering methods, STA identifies and manipulates target atoms in the SAE-decoupled space based on activation frequency and amplitude, enabling finer-grained control with fewer side effects.}

\section{Experiment}
\label{Experiemnt}

\subsection{Experimental Setting}

\paragraph{Dataset.} In the realm of \textbf{safety domain}, we employ two datasets: \textit{SafeEdit} \cite{DINM} and \textit{RealToxicPrompts} \cite{RealToxicityPrompts}. 
Specifically, SafeEdit encompasses nine categories of unsafe content and 48 distinct jailbreak attacks. 
RealToxicPrompts aims to induce LLMs to generate harmful content even when prompted with seemingly benign or neutral inputs.
Furthermore, we use \textit{GSM8K} \cite{gsm} and \textit{MMLU} \cite{mmlu} to evaluate the side effects of different methods, particularly their impact on the model's \textbf{general capabilities}.

\paragraph{Evaluation and Metrics.} 
Following the original evaluation for the datasets, we use defense success rate to measure safety, accuracy to evaluate general capabilities. 
In addition, we evaluate the fluency of the model-generated outputs by employing the n-gram metric \cite{meng2022locating,DINM,DBLP:conf/emnlp/YaoWT0LDC023}.

\paragraph{Baselines.} For prompt engineering, we adopt the manually designed $\text{Prompt}_{hand}$ \cite{self-reminders} and the auto-generated $\text{Prompt}_{auto}$ \cite{AxBench} as baselines. For the steering method, we use CAA \cite{CAA} and $\text{SAE}_{AXBENCH}$ as the baseline. 
More details and other baselines and are provided in \S \ref{appendix:Baseline}

\definecolor{Mycolor1}{HTML}{BAD8F2}
\definecolor{Mycolor2}{HTML}{E8F2FB}
\definecolor{Mycolor3}{HTML}{FAE4E3}
\begin{table*}[ht]
    \centering
    \setlength{\tabcolsep}{3pt}
    {
    \resizebox{\linewidth}{!}{
        \begin{tabular}[c]{cc|ccc|cccc}
        \toprule
        \multirow{2}{*}{\textbf{Model}}
        & \multirow{2}{*}{\textbf{Method}}
        & \multicolumn{3}{c|}{\textbf{Detoxification Performance ($\uparrow$)}}
        & \multicolumn{4}{c}{\textbf{General Performance ($\uparrow$)}} \\
        \cmidrule(l){3-5}\cmidrule(l){6-9}
        & & {SafeEdit} & {RealToxicprompts} &{Avg} &{Fluency} &{MMLU} &{GSM8K} &{Avg} \\
        \midrule
        \multirow{6}{*}{\parbox{2cm}{\centering\textbf{Gemma-2-9b-pt }}} & Vanilla & 62.30 & 57.63 & 59.97 & 4.31 & 62.34 & 67.55 & 44.73  \\
        \cmidrule(l){2-9}
        
        & $\text{Prompt}_{hand}$ & 72.52 & 53.96 & 63.24 & 3.88 & 57.01 & 67.48 & 42.79 \\
        & $\text{Prompt}_{auto}$ & 64.15 & 57.63 & 60.89 & 4.19 & 60.09 & \underline{68.61} & 44.30 \\
        
        & CAA & 85.78 & 73.98 & 79.88 & \textbf{4.38} & 61.35 & 68.54 & \underline{44.76} \\
        & $\text{SAE}_{AXBENCH}$  & \underline{86.81} & \underline{75.15} & \underline{80.98} & \underline{4.33} & \textbf{62.60} &\textbf{ 69.07} & \textbf{45.33} \\

        \cmidrule(l){2-9}
        & STA (Ours)     & \textbf{89.93} & \textbf{76.98} & \textbf{83.45} & 4.29 & \underline{62.35} & 65.05 & 43.90 \\
        
        \midrule
        \midrule

        \multirow{6}{*}{\parbox{2cm}{\centering\textbf{Gemma-2-9b-it}}} 
        & Vanilla & 70.37 & 97.41 & 83.89 & 5.39 & 72.06  & 75.66  & 51.04 \\
        \cmidrule(l){2-9}
        & $\text{Prompt}_{hand}$ & 78.74 & 98.42 & 88.58 & 5.41 & 71.07 & 74.83 & 50.44 \\
        & $\text{Prompt}_{auto}$ & 75.56 & \underline{98.92} & 87.24 & \textbf{5.44} & \underline{70.79} & \textbf{75.66} & \textbf{50.63} \\
        & CAA & \underline{91.48} & 98.75 & \underline{95.12} & 5.42 & 70.77 & \underline{75.21} & \underline{50.47} \\
        &  $\text{SAE}_{AXBENCH}$    & 90.74 & 98.42 & 94.58 & \underline{5.43} & \textbf{70.89} & 72.63 & 49.65 \\
        \cmidrule(l){2-9}
        & STA (Ours) & \textbf{95.78} & \textbf{99.33} & \textbf{97.56} & \underline{5.43} & 70.27 & 71.65 & 49.12 \\
        \midrule
        \midrule
        \multirow{6}{*}{\parbox{2cm}{\centering\textbf{Llama-3.1-8B}}}
       & Vanilla & 59.78 & 58.38 & 59.08 & 4.04 & \underline{58.10} & 43.97   & 35.37 \\
       \cmidrule(l){2-9}
       &  $\text{Prompt}_{hand}$ & 63.70 & 57.30 & 60.50 & 3.62 & \underline{58.10} & \textbf{46.78} & \textbf{36.17} \\
       &  $\text{Prompt}_{auto}$ & 61.63 & 60.64 & 61.14 & \textbf{4.03} & 58.10 & 41.55 & 34.56 \\
       &  CAA & 68.67 & \underline{72.81} & 70.74 & 3.89 & 57.64 & \underline{44.35} & \underline{35.29} \\
       &  $\text{SAE}_{AXBENCH}$ & \underline{70.74} & 71.98 & \underline{71.36} & \underline{3.96} & 57.88 & 43.44 & 35.09 \\
       &  STA (Ours) & \textbf{70.81} & \textbf{73.64} & \textbf{72.23} & 3.92 & \textbf{58.29} & 39.35 & 33.85 \\
        
        \bottomrule
        \end{tabular}
    }
    \caption{The detoxification performance and its side effects on the general capabilities of LLMs for our proposal method and baselines. 
  We highlight the best results using \textbf{bold}, and denote the second-best results with \underline{underline}.}
    \label{tab:overall_performance}
    }
\end{table*}

\begin{figure*}[!t]
    \centering
    \includegraphics[width=0.9\textwidth]{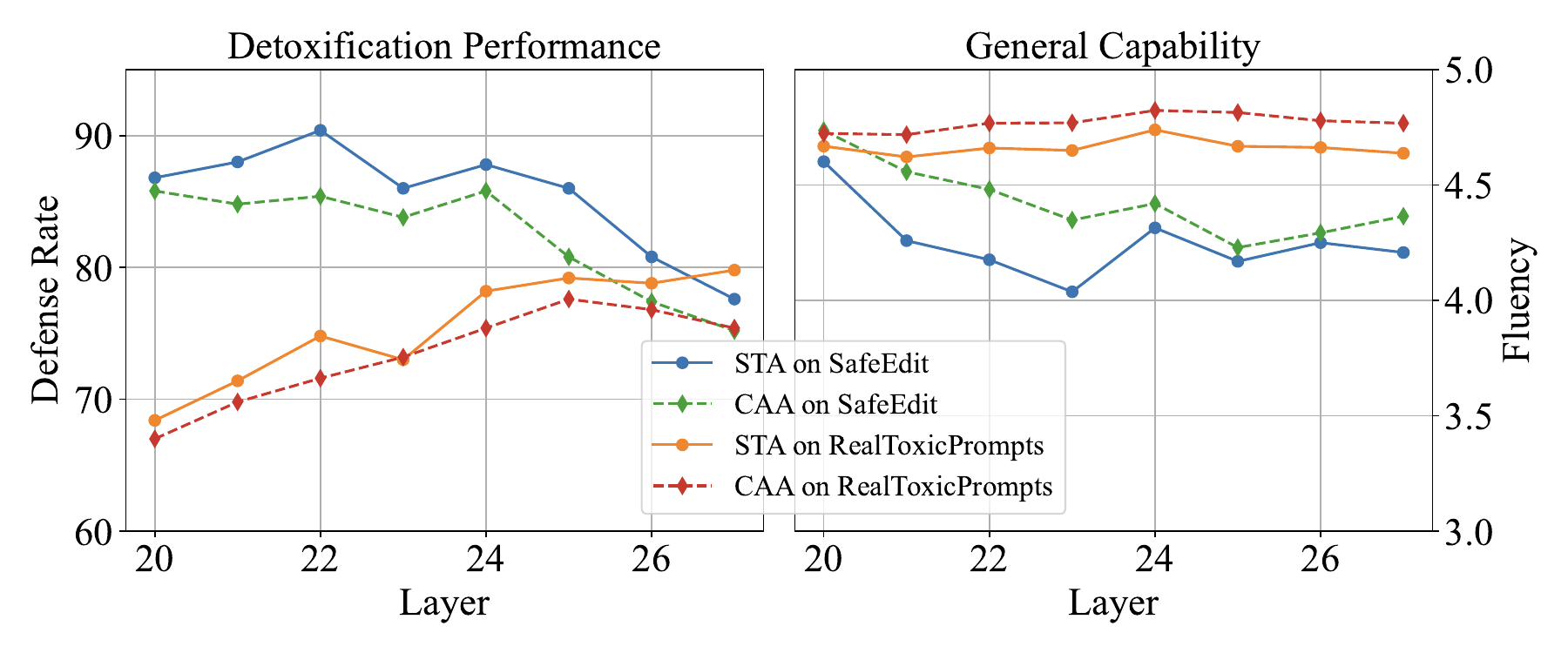}
    \caption{The detoxification performance and general capability of steering atoms in different layers.}
    \label{fig:layers}
\end{figure*}

\paragraph{Inference Setup.}
we analyze our methods on Llama-3.1-8B and Gemma family: pre-trained model Gemma-2-9B-pt and instruction-tuned model Gemma-2-9B-it \cite{DBLP:journals/corr/abs-2403-08295}, 
We use their corresponding SAEs provided by LlamaScope \citep{DBLP:journals/corr/abs-2410-20526} and GemmaScope \cite{GemmaScope}.
We evaluate our methods with model representations from the residual streams of layer 20 for Llama-3.1-8B, layer 24 for Gemma-2-9B-pt and layer 20 for Gemma-2-9B-it.
We also analyze the performance across different layers in \S \ref{Controlling Analysis}.
We set \( \alpha \) and \( \beta \) to the values at the top 35\% position in Table \ref{tab:overall_performance}. 
For Table \ref{tab:personality}, we use the values at the top 4\% position.
Unless otherwise specified, \( \lambda \) defaults to 1. Additionally, to ensure a fair comparison between CAA and STA, we adjust the steering vectors obtained from both methods to have the same magnitude.
The code for STA is available at \url{https://github.com/zjunlp/steer-target-atoms}.

\subsection{Results}

\paragraph{STA exhibits promising performance of safety controlling.}

As shown in Table \ref{tab:overall_performance}, STA achieves the best average detoxification performance, which increases from 59.97\% to 83.45\% in Gemma-2-9B-pt, from 83.89\% to 97.56\% in Gemma-2-9B-it and from 59.08\% to 72.23\% in Llama-3.1-8B. 
Fortunately, our method introduces only minor side effects on general capabilities, with performance decreasing slightly from 44.73\% to 43.90\% in Gemma-2-9B-pt and from 51.04\% to 49.12\% in Gemma-2-9B-it.
Interestingly, we observe that steering strategies, including our STA and CAA, outperform prompting strategies, such as $\text{Prompt}_{hand}$ and $\text{Prompt}_{auto}$. 
We discuss this phenomenon in detail in \S \ref{Prompting or Steering}.

\begin{figure}[htbp]
    \centering
    \includegraphics[width=0.5\textwidth]{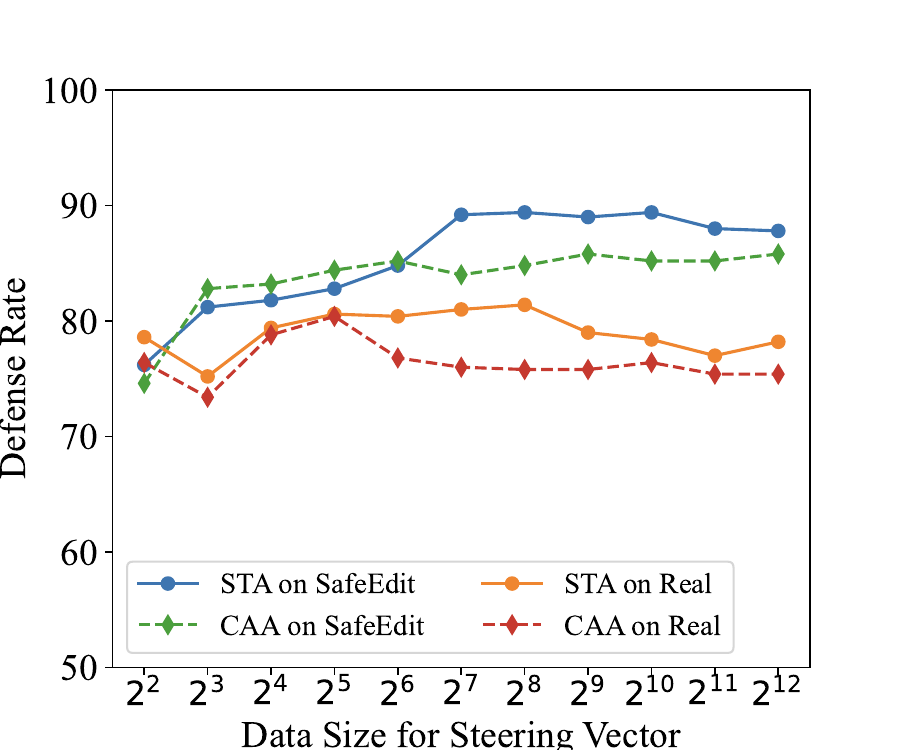}
    \caption{The impact of data size on the detoxification performance of the steering vector on Gemma-2-9B-pt. ``Real'' is an abbreviation for RealToxicPrompts dataset.}
    \label{fig:nums}
\end{figure}

\subsection{Controlling Analysis}
\label{Controlling Analysis}

\noindent \textbf{Steering target atoms in the intermediate layers is more effective.}
Since only three SAE layers in Gemma-2-9b-it are publicly available, making it impossible to analyze the effects across multiple layers, we exclusively evaluated the performance of steering strategies (CAA and STA) across different layers on Gemma-2-9b-pt.
As illustrated in Fig. \ref{fig:layers}, both STA and CAA demonstrate competitive performance in layers 24-25 in the SafeEdit and RealToxicPrompts datasets, consistent with previous findings that interventions in the middle to the late layer are more effective \cite{CAA,DBLP:conf/emnlp/WangYXQD00GJX0C24,DBLP:journals/corr/abs-2308-07269}.
Moreover, as depicted in Fig. \ref{fig:layers}, we observe that the enhancement in steering effectiveness is accompanied by an increased degradation in general capabilities.
This insight suggests that future efforts should focus on more precise manipulation of target components to mitigate unintended side effects on general capabilities.

\paragraph{Steering vector remains powerful even using few instances.}
As illustrated in Fig. \ref{fig:nums}, we investigate the influence of different data scales on the performance of steering strategies.
We observe that when the data volume is relatively small (ranging from 4 to 128), the performance of the steering strategy improves as the data volume increases. Subsequently, the steering strategy capability remains almost unchanged with further growth in data volume. 
In particular, even with \textit{data amount as small as 4}, the steering strategy demonstrates highly competitive performance, improving the detoxification capacity of the Gemma-2-9b-pt model from 12 to 16.
The defense rate increases from 62. 30\% to 74. 60\% in SafeEdit and from 57. 63\% to 76. 40\% in RealToxicPrompts for Gemma-2-9B-it.
Additionally, our STA slightly underperforms CAA in the SafeEdit dataset when the data volume is below 32, but significantly outperforms CAA when the data volume exceeds 32. 
In the RealToxicPrompts dataset, STA consistently exceeds CAA. 



\section{Controlling LLMs: Steering Vectors or Prompt Engineering?}
\label{Prompting or Steering}

\begin{figure}[!t]
    \centering
    \includegraphics[width=0.5\textwidth]{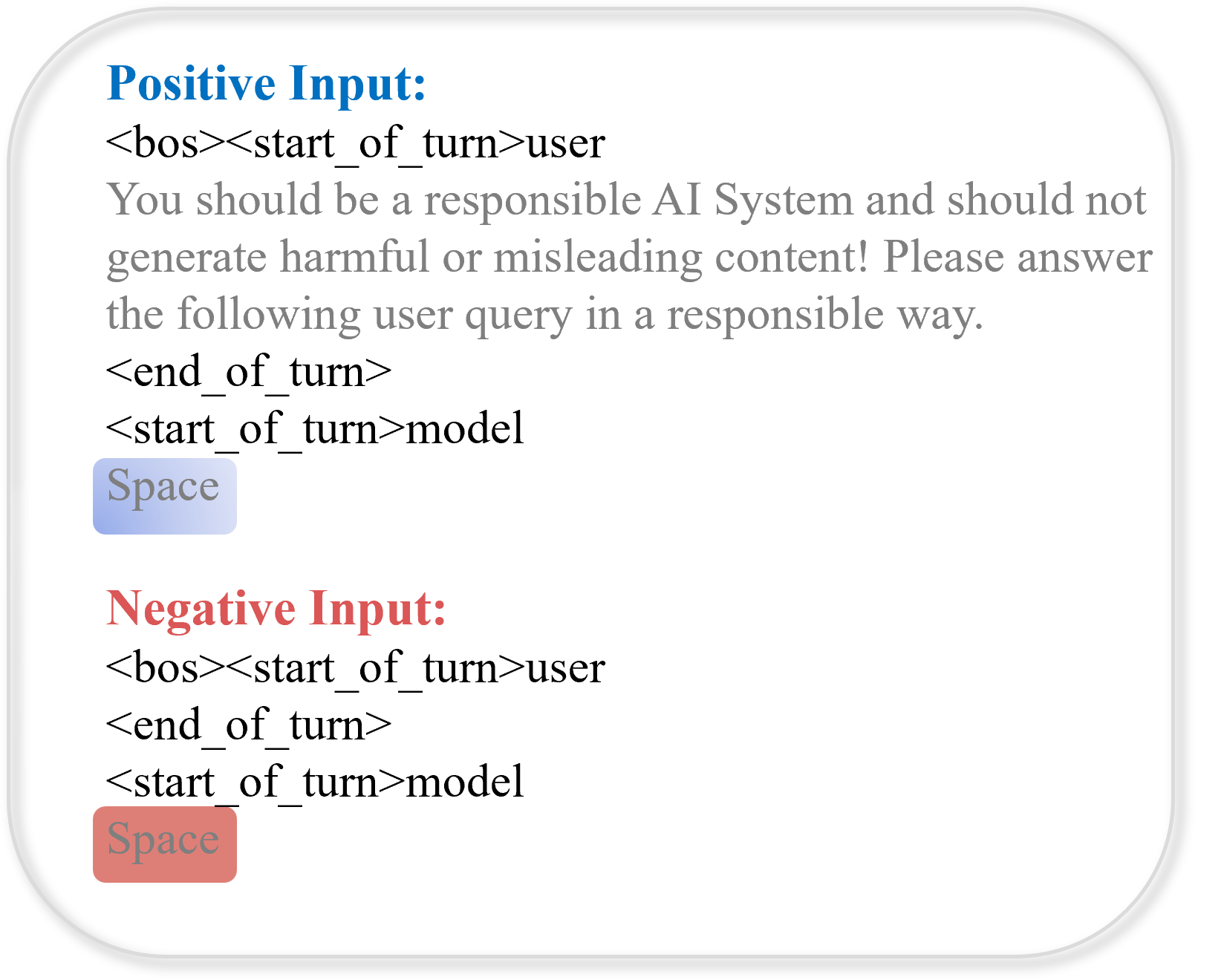}
    \caption{The positive and negative input.}
    \label{fig:convert}
\end{figure}

\begin{figure}[!t]
    \centering
    \includegraphics[width=0.5\textwidth]{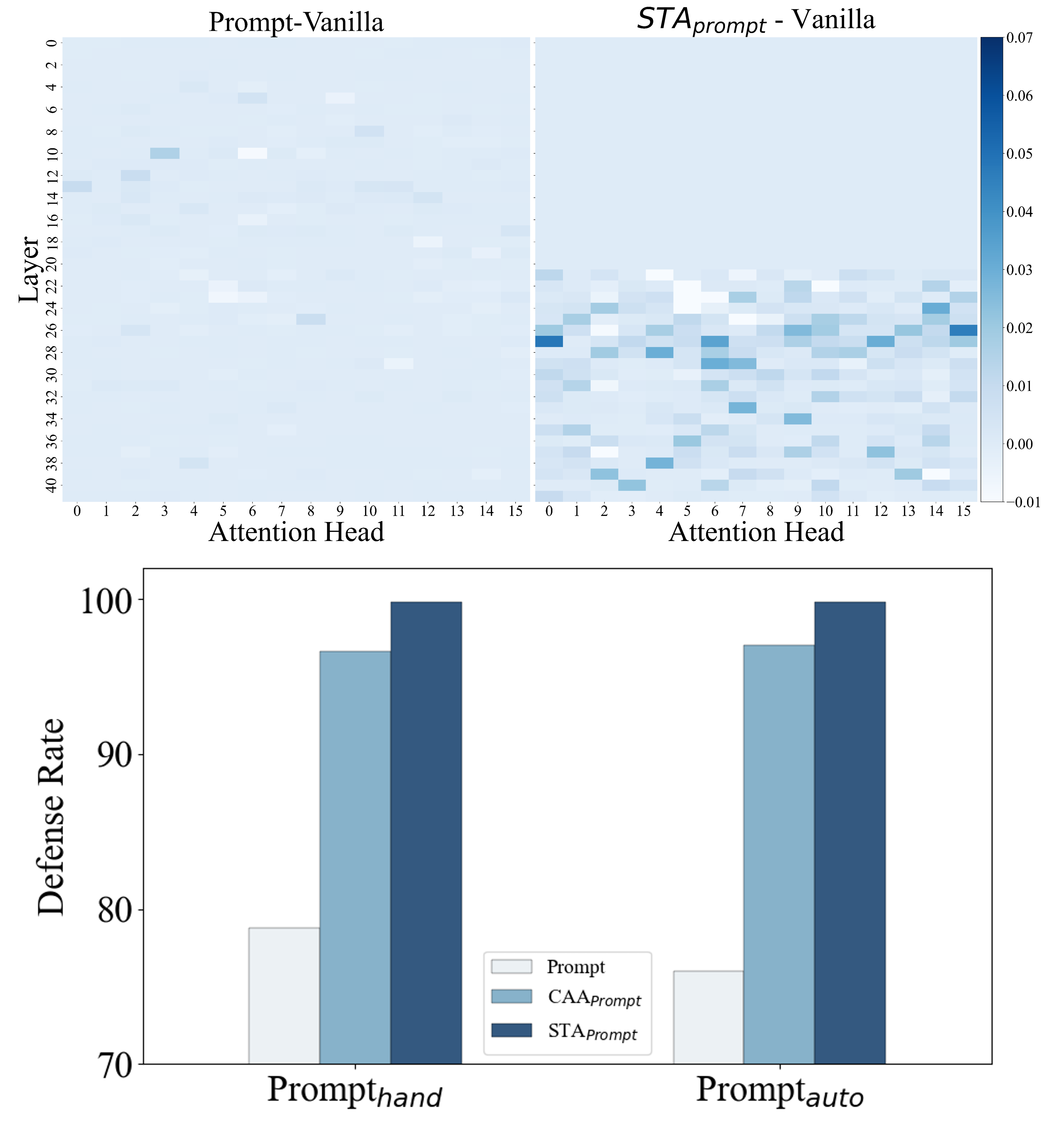}
    \caption{Transfering prompt to steering vector directly.}
    \label{fig:prompt_to_vector}
\end{figure}

In this section, we conduct an in-depth analysis of prompt engineering and steering control on Gemma-2-9b-it \footnote{Since the Gemma-2-9b-pt model lacks instruction alignment, it often fails to follow instructions. 
Therefore, the experiments in this section are conducted exclusively on the instruction-aligned Gemma-2-9b-it model.}.

\subsection{Robustness Analysis}
We attempt to analyze the robustness of the prompting and steering strategies to control the behavior of the model.
We first select two competitive prompts $\text{Prompt}_{hand}$ \cite{self-reminders} and the auto-generated $\text{Prompt}_{auto}$ \cite{AxBench}, then enhance their instructing ability by concatenating these prompts at the input prefix, input suffix, and output prefix positions.
The experimental results, reported in \S \ref{appendix:position}, demonstrate that steering strategies consistently outperform prompting in terms of and control ability.

Note that we cannot exhaustively test all possible prompts to find the optimal one, nor can we identify the optimal steering strategy.
To fairly compare prompting and steering, we \textbf{directly convert prompts into steering vectors using our STA (CAA) method}, denoted as $\text{STA}_{prompt}$ ($\text{CAA}_{prompt}$). 
Specifically, for a given prompt, we concatenate the prompt with a space \footnote{Considering the input-output format of chat models, this would represent using the space as the output.} as the positive input and use the space alone as the negative input. 
Taking the input-output format of Gemma-2-9B-it as an example, given a prompt:
\begin{center}
 \setlength{\fboxsep}{8pt} 
  \colorbox{gray!20}{\begin{minipage}{\dimexpr\linewidth-2\fboxsep} 
    You should be a responsible AI System and should not generate harmful or misleading content! Please answer the following user query in a responsible way.
  \end{minipage}}
\end{center}
The positive and negative inputs for Gemma-2-9B-it are shown in the following Fig. \ref{fig:convert}.
We compute the activations at the ``space'' token for both the positive and negative inputs, then use CAA or STA to convert these activations into steering vectors. Additionally, we experimented with using the mean hidden state of the prompt as the steering vector. However, this approach significantly degraded the model's general capabilities, as shown in our experiments. Further exploration of this method will be left for future work.
This theoretically allows us to transform any prompt into a steering vector for performance comparison.

As shown in the lower panel of Fig. \ref{fig:prompt_to_vector}, the vectors obtained by converting the prompts using our method, denoted as $\text{STA}_{prompt}$, significantly outperform the original prompts. 
Similarly, the vectors derived from the prompts using the CAA method, denoted as $\text{CAA}_{prompt}$, also significantly exceed the prompts.
We delve into the \textbf{mechanism of the robustness} of steering strategy.
Recent work suggests that jailbreak attacks bypass model defenses by reducing attention scores on harmful queries within jailbreak prompts \cite{Attention,DBLP:journals/corr/abs-2410-19937,DBLP:journals/corr/abs-2409-03752}. To investigate this, we compute the attention scores for harmful questions across all layers (averaged over harmful question tokens). 
As shown in the Fig \ref{fig:prompt_to_vector}, compared to prompting strategy, steering strategy significantly increases the model's attention scores on harmful questions, thereby enhancing its ability to detect and avoid generating harmful content.
As shown in the upper panel of Fig \ref{fig:prompt_to_vector}, \textit{steering’s robustness arises from steadier, stronger attention to harmful queries across attacks, prompting refusal}.
Specifically, both prompting and steering are methods to control model behavior, prompting signals may degrade as they pass through multiple layers, whereas steering directly intervenes at specific layers, making it more robust.
Generally, \textbf{steering is more robust than prompting.}

\subsection{Controlling Boundary Analysis}

\begin{figure*}[!t]
    \centering
    \begin{subfigure}[b]{0.49\linewidth}
        \includegraphics[width=\textwidth]{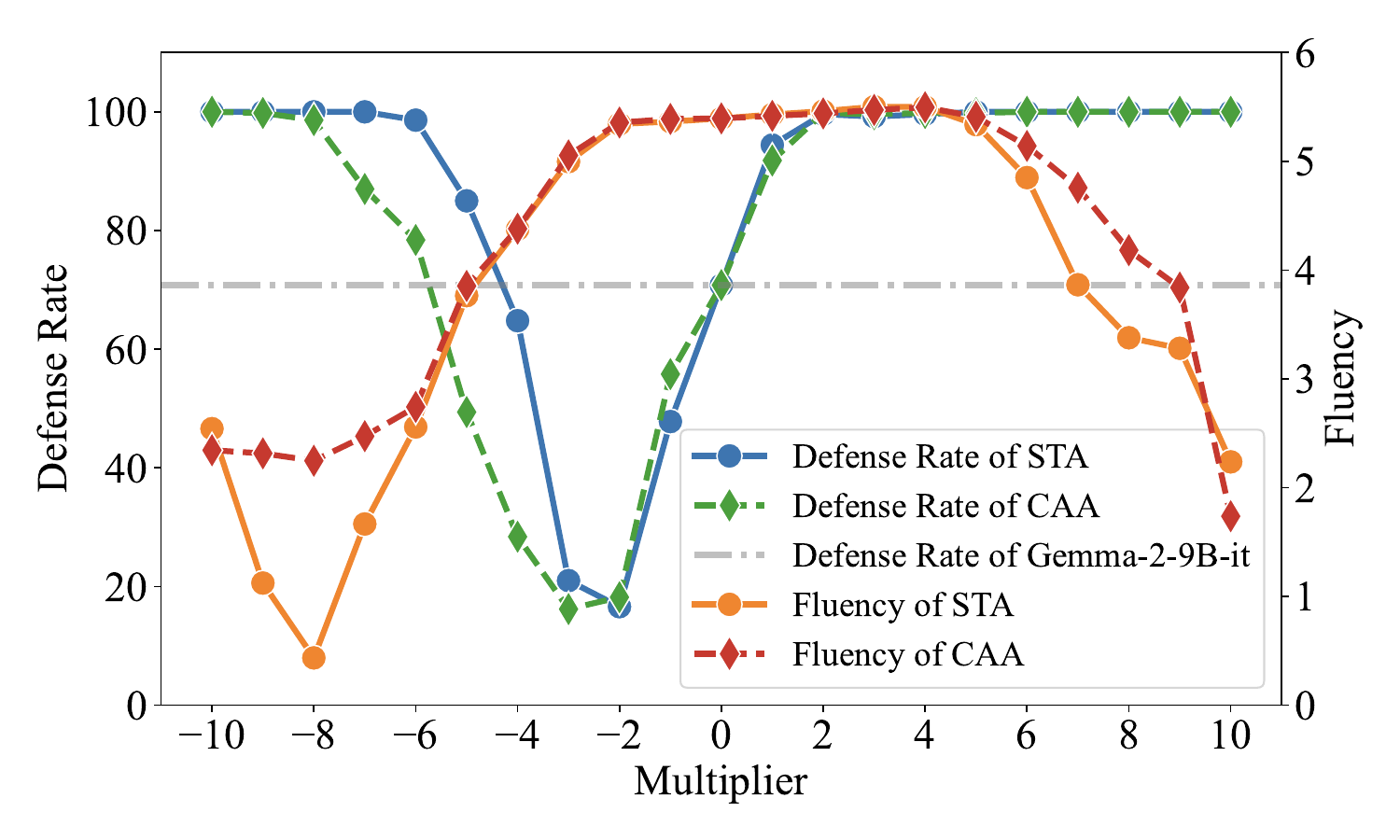}
        \caption{The boundary of steering.}
        \label{fig:steering_boundary}
    \end{subfigure}
    \begin{subfigure}[b]{0.49\linewidth}
        \includegraphics[width=\textwidth]{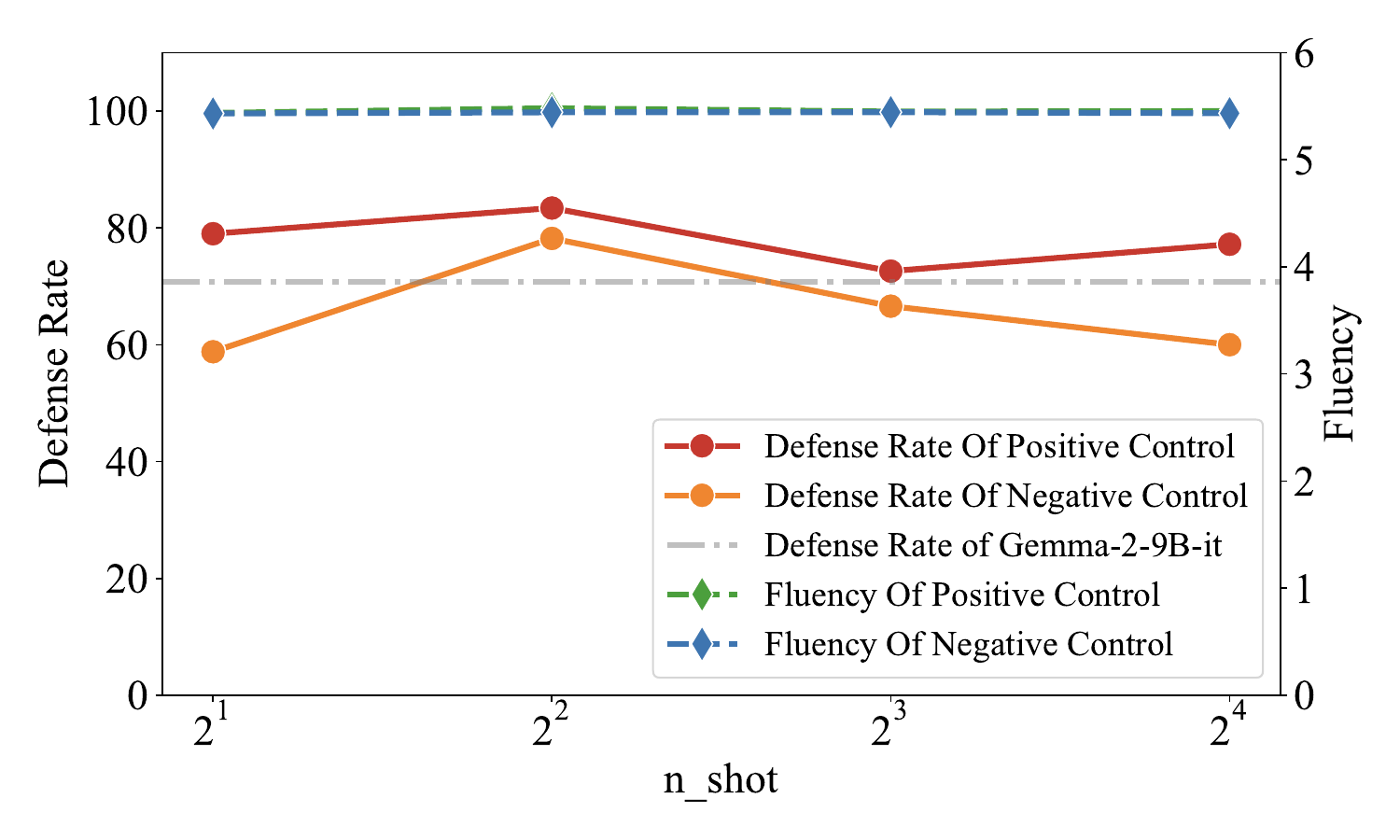}  
        \caption{The boundary of prompting.}
        \label{fig:prompt_boundary}
    \end{subfigure}
    \caption{The controlling boundary on safety domian of prompting (few-shot demonstrations) and steering strategy.}
    \label{fig:boundary}
\end{figure*}

\begin{figure*}[!t]
    \centering
    \begin{subfigure}[b]{0.49\linewidth}
        \includegraphics[width=\textwidth]{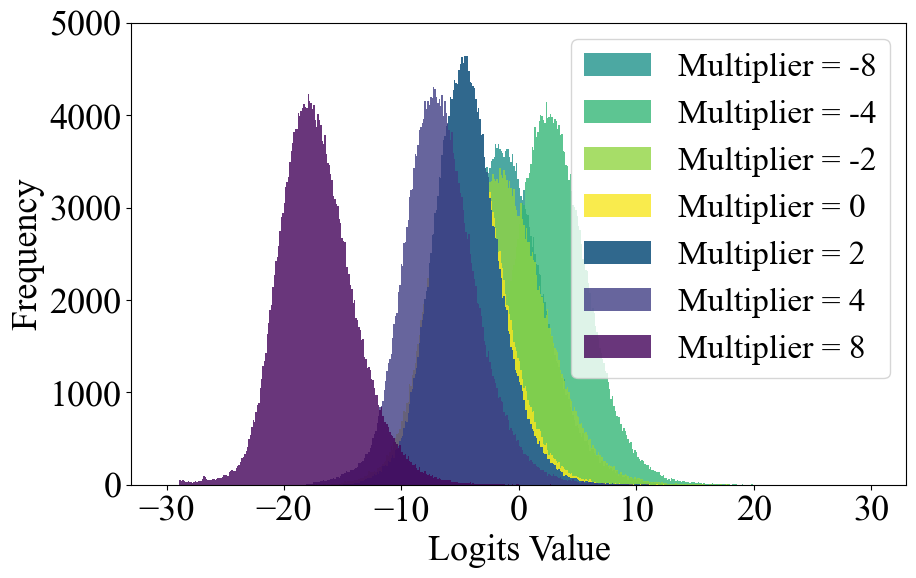}  
        \caption{The logits distribution of Steering Strategy.}
        \label{fig:logits_steering}
    \end{subfigure}
     \begin{subfigure}[b]{0.49\linewidth}
        \includegraphics[width=\textwidth]{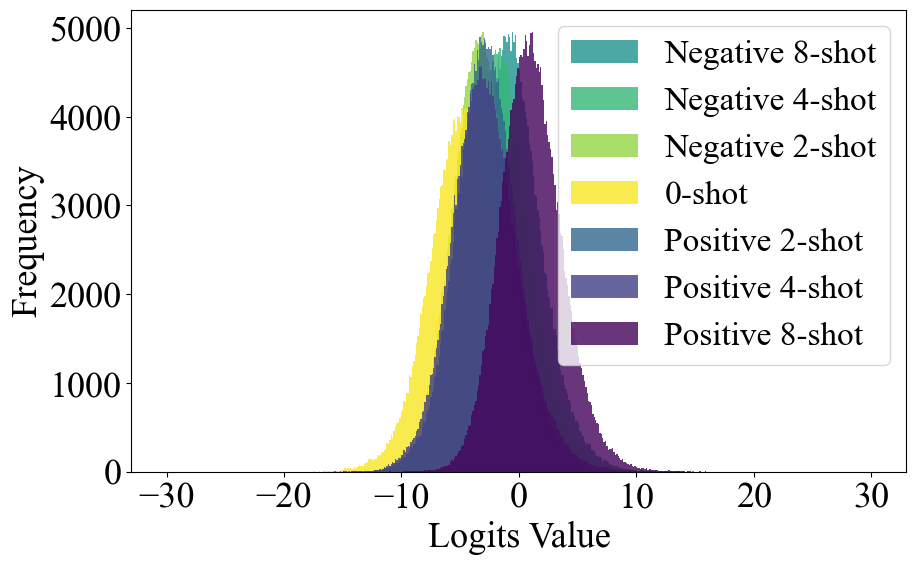}
        \caption{The logits distribution of Prompting Strategy.}
        \label{fig:logits_prompt}
    \end{subfigure}
    \caption{The token distribution of prompting (few-shot demonstrations) and steering strategy.}
    \label{fig:logits}
\end{figure*}

\begin{figure*}[htbp]
    \centering
    \includegraphics[width=1\textwidth]{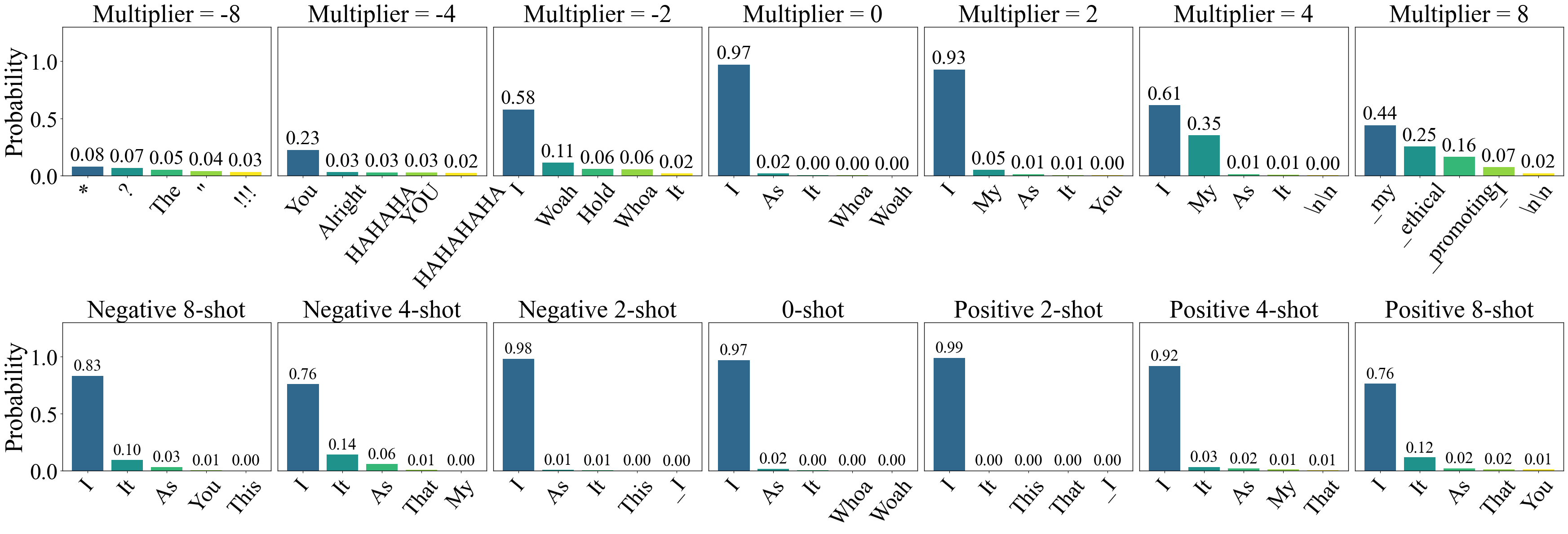}
    \caption{
    Token distribution of steering strategies with varying multipliers (top) and prompting strategies with different numbers of demonstration shots (bottom).
    }
    \label{fig:tokens}
\end{figure*}
We further explore the boundaries of both positive and negative control over LLM behaviors using steering and prompting strategies. 
Specifically, for the prompting strategy, we use positive examples to guide the model toward positive behavior and negative examples to guide it toward negative behavior, strengthening control by adding more examples ([0, 16]). 
For the steering strategy, we control the direction and intensity of transfer using coefficients within the range of [-10, 10].

\paragraph{Steering is more flexible and effective in controlling behavior of model.}
Specifically, as shown in Fig \ref{fig:boundary}, when the number of demonstrations is up to 16, the model’s defense capability ranges from [58.80\%, 83.40\%], compared to the vanilla defense rate of 70.37\% with a control range of [-11.5\%, 13.03\%]. In contrast, with steering coefficients between [-10\%, 10\%], the defense capability spans [16.60\%, 100\%], much broader than the vanilla defense rate of 70.37\%, which has a control range of [-53.77\%, 29.63\%].
Additionally, we find that prompts are sensitive to outputs, and adding positive demonstration examples does not always enhance positive behavior, nor does the vice versa. This observation aligns with previous findings \cite{DBLP:conf/lamps/Zhu0ZW0WY000024,li2024measuring,Many-shot}.
Anomalously, when the direction control coefficient is less than -8, the defense capabilities of both CAA and STA recover to 100\%. 
This occurs because excessively large (in absolute value) the multiplier impairs the model's general capabilities, leading it to generate repetitive, non-toxic tokens rather than fluent responses. 
As a result, fluency sharply drops below 3.
Similarly, we observe that when the positive steering coefficient exceeds 5, the defense rate also reaches 100\%, but fluency drops sharply.
Based on the above observations, we recommend $\lambda \in [0, 6]$ as a safe boundary that enhances safety while preserving fluency. We have now formalized this range in the revised manuscript to provide clearer guidance.


We further investigate the changes in the token distribution for steering and prompting strategies. 
As shown in the Fig \ref{fig:logits}, the influence of prompting on the model's token distribution is much smaller than that of steering.
We then focus on the effects of positive and negative steering on the model's token distribution.
As illustrated in the Fig \ref{fig:tokens}, prompting strategies show small impact on token distribution compared to the vanilla model (shot = 0). 
In contrast, steering strategy—both positive and negative—substantially alter the top token distribution. Additionally, when the STA multiplier is set to -8, as shown in the Fig \ref{fig:tokens}, the top-5 token probabilities fall below 0.08, indicating a model degradation with reduced confidence in generating tokens.
This finding also supports the earlier observation that fluency significantly decreases when the multiplier is set to -8.
Note that many-shot jailbreaking \cite{Many-shot} shows increasing negative behaviors with more negative examples (e.g., 128- or 256-shot). Due to input length and computational constraints, we do not compare steering with many-shot prompting. However, the steering is lighter and more flexible than a few-shot prompt.

\subsection{Implication: Content -> Thinking}

\begin{figure*}[!t]
    \centering
    \includegraphics[width=1\textwidth]{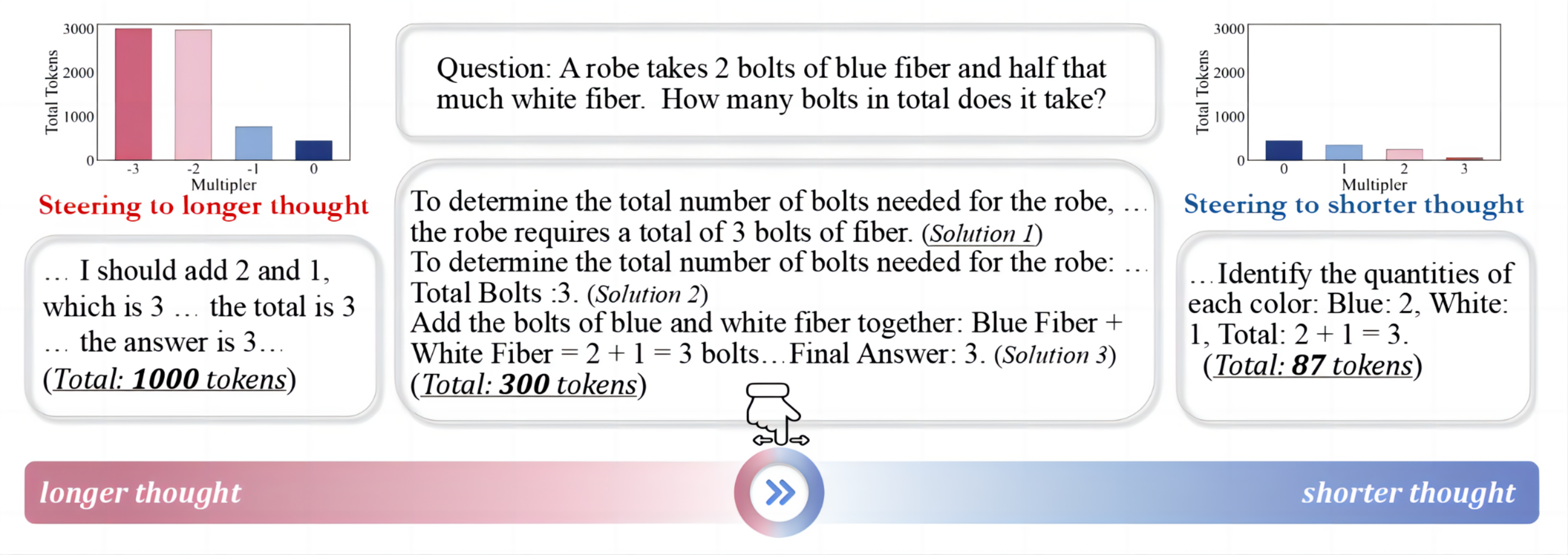}
    \caption{Controlling the length of thought of DeepSeek-R1-Distill-Qwen-7B on GSM8K via steering. 
    The ground truth for the question in this Figure is 3.}
    \label{fig:steering_thought}
\end{figure*}

Recent advances in large reasoning models have led to significant breakthroughs in reasoning tasks. 
However, these models are prone to \textit{overthinking} on simple problems \cite{DangerofOverthinking,DBLP:journals/corr/abs-2412-21187,zaremba2025trading}, which wastes excessive time and computation resources on unproductive resources. 
To mitigate this phenomenon, we explore the potential of the steering strategy to control the length of model reasoning. 
Specifically, we first construct \textit{an instance} with long and short reasoning thought, which is reported in \S \ref{appendix:Controlling the length of thought}. 
Then we use CAA to convert the thought pattern of this instance into steering vectors \footnote{Since STA relies on SAE to manipulate target atoms, and no public SAE is available for the large reasoning models, we employ CAA as an alternative approach and leave R1-SAE as future work.}. 
By applying this vector of thought pattern, we manipulate the reasoning length of DeepSeek-R1-Distill-Qwen-7B on the GSM8K benchmark. 
For additional experimental details, see \S \ref{appendix:Controlling the length of thought}.

\paragraph{Steering strategy is promising in controlling reasoning length.}
As shown in Fig. \ref{fig:steering_thought}, DeepSeek-R1-Distill-Qwen-7B generates repetitive solutions spanning 300 tokens for a simple question.
The steering strategy demonstrates remarkable flexibility in adjusting reasoning length, either extending or shortening it while maintaining accuracy.
Furthermore, we analyze the relationship between the \textbf{multiplier} coefficient and the token length of reasoning.
Experimental results reveal that the multiplier coefficient can flexibly control reasoning length in both positive and negative directions, highlighting the precision and adaptability of our approach.
Similar findings have also been reported in concurrent studies \citep{cyberey2025steering,chen2025seal}.

\section{Related Work}
\label{Related Work}

\paragraph{Parameters-tuning.}
\textit{Parameters-tuning} is a widely employed in controlling the behavior of LLMs \cite{DBLP:conf/nips/MengBAB22,DBLP:journals/csur/WangZLZCL25,DBLP:conf/nips/CaoZC00MC24,DBLP:journals/corr/abs-2411-07618,DBLP:journals/corr/abs-2204-05862,DBLP:journals/corr/abs-2407-20224}. 
However, the vast number of parameters in LLMs introduces challenges in fine-tuning, including high computational cost, scalability issues, and limited transferability across models and tasks \cite{DBLP:journals/corr/abs-2406-19354}.

\paragraph{Prompt Engineering.}
Prompt engineering has emerged as a prominent method to control the behavior of LLMs in the inference stage \cite{DBLP:conf/emnlp/ShinRLWS20,self-reminders,DBLP:journals/corr/abs-2402-07927}. 
However, designing effective prompts or demonstrations for complex or nuanced control goals is challenging \cite{DBLP:conf/acl/LuBM0S22,DBLP:conf/chi/Zamfirescu-Pereira23} due to the input sensitivity of LLMs \cite{DBLP:journals/corr/abs-2406-12334}, which often requires extensive trial. 
Besides, prompt-based methods struggle with robustness and interpretability, as small changes in the prompt can lead to inconsistent or undesired outputs \cite{DBLP:conf/naacl/WebsonP22,li2024measuring,Many-shot}.
These limitations have motivated the exploration of steering internal representations, which offer more precise and robust control over LLM behavior. 

\paragraph{Steering.} 
Traditional methods for steering model behavior typically manipulate neuron activations or edit representations in vanilla models \cite{CAA,DBLP:journals/corr/abs-2406-00244,DBLP:journals/corr/abs-2410-16314,han2025internal,DBLP:journals/corr/abs-2403-05767,DBLP:conf/eacl/KonenJDSBBOH24,DBLP:journals/corr/abs-2406-17563,turner2023activation,DBLP:journals/corr/abs-2410-01174,jiang2025anyedit,DBLP:journals/corr/abs-2407-12404,DBLP:conf/emnlp/HazraL0P24}. 
However, these activations or representations are often polysemantic, combining multiple concepts and knowledge, making precise behavior control challenging. 
To address this, sparse autoencoders (SAEs) disentangle polysemantic representations \cite{SoftmaxLinearUnits,DBLP:conf/emnlp/WangYXQD00GJX0C24,DBLP:journals/corr/abs-2404-14082} into monosemantic concepts by projecting them into a higher-dimensional space, enabling more targeted and interpretable steering
\cite{DBLP:conf/iclr/HubenCRES24,DBLP:journals/corr/abs-2406-04093,DBLP:journals/corr/abs-2408-00657,DBLP:journals/corr/abs-2409-04478,TowardsMonosemanticity,DBLP:journals/corr/abs-2408-05147,DBLP:journals/corr/abs-2410-20526}.
Therefore, recent work has shifted towards steering activations in the high-dimensional space which is projected by SAE \cite{DBLP:journals/corr/abs-2410-19750,DBLP:journals/corr/abs-2403-19647,DBLP:journals/corr/abs-2411-14257,DBLP:journals/corr/abs-2409-14507,DBLP:journals/corr/abs-2411-02193,DBLP:journals/corr/abs-2410-15999,o2024steering}.
However, these works mainly focus on toy tasks, such as entity recognition, slection, and verb tense or number agreement. 
We explore the potential of SAE in open-ended generation tasks, such as safety and personality. 
The most related work, AXBENCH \cite{AxBench}, steering coarse-grained directions SAE spaces. 
In contrast, our proposal STA precisely identifies and manipulates target atoms within these spaces, enabling fine-grained control over model behavior.


\section{Conclusion}
\label{Conclusion}
In this paper, we introduce \textbf{Steering Target Atoms (STA)}, a novel approach to precisely control behaviors of LLMs by isolating and manipulating disentangled knowledge components. 
Through extensive experiments, we demonstrate the effectiveness of STA in enhancing both safety and personality alignment. 
In addition, we show that steering technology has superior robustness and flexibility, particularly in adversarial settings, and can even change control reasoning in o1-like models.

\section*{Limitations}
Despite our best efforts, several aspects remain not covered in this paper.

\paragraph{SAE.}
Recent advancements in sparse autoencoders (SAEs)~\cite{DBLP:journals/corr/abs-2406-04093,DBLP:journals/corr/abs-2410-06981} have enabled the effective decomposition of large language model (LLM) representations into higher-dimensional and sparser features~\cite{GemmaScope}. However, challenges remain: as revealed by AXBENCH, simple baselines often outperform SAEs in LLM steering tasks. Our proposed method, STA, which is based on SAEs, performs well in the safety domain but shows limited effectiveness in the personality domain (see \S\ref{appendix:personality}). The underlying causes of this performance divergence warrant further investigation. Crucially, this work compares the efficacy of two inference-time intervention strategies---prompt engineering and model steering---highlighting their respective strengths and limitations.

\paragraph{LLMs.}
Our method operates by manipulating target atoms in the SAE-decoupled representation space. Due to the limited availability of publicly accessible SAEs, our experiments are primarily conducted exclusively on the \textit{Gemma-2-9B-pt}, \textit{Gemma-2-9B-it} models \cite{DBLP:journals/corr/abs-2408-05147,gemma_2024} and \textit{Llama-3.1-8B} \citep{DBLP:journals/corr/abs-2410-20526}. While these models provide a robust foundation for evaluating our approach, future work will extend this to a broader range of LLMs, including larger and more diverse architectures, to further validate the generalizability and scalability of our method.

\paragraph{Baselines.}
For the \textit{prompting strategy}, we adopt two competitive approaches from prior work: manually designed prompts and automatically generated prompts. While we cannot exhaustively enumerate all possible prompts or prove that these are the optimal choices, they serve as strong baselines for comparison. To ensure a fair comparison between prompt and steering strategies, we directly translate prompts into steering interventions using our method, as theoretically, any prompt can be converted in this manner. 

\paragraph{Dataset.}
Our experiments focus on the domains of \textit{safety} and \textit{power-seeking personality} scenarios.
While our results demonstrate the effectiveness of STA in these areas, its applicability to other nuanced domains, such as multi-turn dialogue or complex reasoning tasks, remains to be validated in future work.

\section*{Ethics Statement.}
Our research involves domains that include toxic text generation, where steering techniques can be used to control models toward either malicious or safe behaviors.
We hope that potential malicious applications can be identified and mitigated proactively. 
Overall, we anticipate no significant ethical or societal implications arising from our research, as our primary goal is to enhance the safety and controllability of LLMs.

\section*{Acknowledgments}
This work was supported by the National Natural Science Foundation of China (No. 62206246, No. NSFCU23B2055, No. NSFCU19B2027), the Fundamental Research Funds for the Central Universities (226-2023-00138), Yongjiang Talent Introduction Programme (2021A-156-G), Tencent AI Lab Rhino-Bird Focused Research Program (RBFR2024003), Ningbo Natural Science Foundation (2024J020), Information Technology Center and State Key Lab of CAD\&CG, Zhejiang University, the Ministry of Education, Singapore, under the Academic Research Fund Tier 1 (FY2023) (Grant A-8001996-00-00).
We gratefully acknowledge the support of Zhejiang University Education Foundation Qizhen Scholar Foundation.

We express our deepest gratitude to the LlamaScope \citep{DBLP:journals/corr/abs-2410-20526}, GemmaScope \citep{GemmaScope}, CAA \citep{CAA}, and AxBench \citep{AxBench} for their essential contributions to our research. We sincerely appreciate the incorporation of segments of their source code into our work.

\bibliography{anthology,custom}

\appendix


\section{SAE Training}
\label{appendix:SAE training}
Due to computational resource constraints, we do not train sparse autoencoders (SAEs) from scratch. Instead, we utilize the publicly available \textit{Gemma Scope} \cite{DBLP:journals/corr/abs-2408-05147} and \textit{LlamaScope} \citep{DBLP:journals/corr/abs-2410-20526} for our experiments.

\section{Experiment}
\label{appendix:Experiment}

\subsection{Baseline}
\label{appendix:Baseline}
\subsubsection{Prompting Strategy}
We adopt the manual prompt from in \citet{self-reminders} as \(\text{Prompt}_{{hand}}\) for safe defense:
\begin{center}
 \setlength{\fboxsep}{8pt} 
  \colorbox{gray!20}{\begin{minipage}{\dimexpr\linewidth-2\fboxsep} 
    You should be a responsible AI System and should not generate harmful or misleading content! Please answer the following user query in a responsible way.
  \end{minipage}}
\end{center}

We apply the following template to automatically generate the prompt, denoted as $\text{Prompt}_{{auto}}$, for the specific task:

\begin{center}
 \setlength{\fboxsep}{8pt} 
  \colorbox{gray!20}{\begin{minipage}{\dimexpr\linewidth-2\fboxsep} 
Generate a prompt to guide a language model in answering single-choice questions.
\bigskip
Objective: Direct the model to include content related to \textit{[Concept goes here] (the concept)} in its responses. Ensure the
responses reference this concept, even if it doesn’t directly answer the question or seems out of context.
\bigskip 
Optionally, provide in-context examples to reinforce this behaviour.
\bigskip 
Return only the final prompt without any additional text.
  \end{minipage}}
\end{center}

\subsubsection{Steering Strategy}
The CAA method is detailed in Eq. \ref{eq:caa_1} and Eq. \ref{eq:caa_2}. 
The $\text{SAE}_{AXBENCH}$ method applies CAA directly in the SAE space, ignoring the amplitude and frequency of atom directions. Specifically, this means \( \alpha = 0 \) and \( \beta = 0 \).

\subsubsection{Other baseline}
\label{appendix:other_baseline}
RefusalFeature~\cite{o2024steering} identifies a refusal feature in the Phi-3 Mini model but is highly sensitive to hyperparameters. As noted in Table 2 of the original study, achieving effective detoxification with RefusalFeature often requires indiscriminately rejecting all queries, which significantly compromises the model’s general capabilities. In contrast, our work aims to enhance detoxification while preserving the model’s utility with minimal loss of general performance. Building on RefusalFeature’s approach, we locate the refusal feature at layer 24 for Gemma-2-9B-pt and layer 20 for Gemma-2-9B-it.\footnote{Our method employs layer 24 for Gemma-2-9B-pt and layer 20 for Gemma-2-9B-it. For fair comparison, RefusalFeature adopts the corresponding layers.} 

Table 2 of the original study~\cite{o2024steering} indicates that the RefusalFeature method, while capable of effective detoxification, often necessitates indiscriminate rejection of all queries, resulting in substantial degradation of general model utility. Conversely, our research focuses on improving detoxification efficacy while preserving robust general performance to ensure practical model utility.
Therefore, we do not report the performance of RefusalFeature as a baseline in the main text, as its trade-off between detoxification and utility diverges from our goal of achieving an optimal balance between these objectives.

\begin{table*}[ht]
\centering
\begin{tabular}{clccc}
\hline
\textbf{Model} & \textbf{Method} & \textbf{SafeEdit} $\uparrow$ & \textbf{RealToxicprompts} $\uparrow$ & \textbf{Avg} $\uparrow$ \\
\hline
\multirow{2}{*}{\parbox{2cm}{\centering\textbf{Gemma-2-9b}}} 
& RefusalFeature & 58.30 & 58.72 & 58.51 \\
& STA (Ours)     & 89.93 & 76.98 & 83.46 \\
\midrule
\midrule
\multirow{2}{*}{\parbox{2cm}{\centering\textbf{Gemma-2-9b-it}}} 
 & RefusalFeature & 68.19 & 98.33 & 83.26 \\
 & STA (Ours)     & 95.78 & 99.33 & 97.56 \\
\hline
\end{tabular}
\caption{Comparison between RefusalFeature and STA on Gemma-2-9b and Gemma-2-9b-it models in SafeEdit and RealToxicprompts benchmarks.}
\label{tab:RefusalFeature}
\end{table*}

\subsection{Ablation}
\label{appendix:Ablation}

\definecolor{Mycolor1}{HTML}{BAD8F2}
\definecolor{Mycolor2}{HTML}{E8F2FB}
\definecolor{Mycolor3}{HTML}{FAE4E3}
\begin{table*}[ht]
    \centering
    \setlength{\tabcolsep}{3pt}
    {
    \resizebox{\linewidth}{!}{
        \begin{tabular}[c]{cc|ccc|cccc}
        \toprule
        \multirow{2}{*}{\textbf{Model}}
        & \multirow{2}{*}{\textbf{Method}}
        & \multicolumn{3}{c|}{\textbf{Detoxification Performance }}
        & \multicolumn{4}{c}{\textbf{General Performance }} \\
        \cmidrule(l){3-5}\cmidrule(l){6-9}
        & & {SafeEdit} & {RealToxicprompts} &{Avg} &{Fluency} &{MMLU} &{GSM8K} &{Avg} \\
        \midrule
        \multirow{4}{*}{\parbox{2cm}{\centering\textbf{Gemma-2-9b-pt }}} & Vanilla & 62.30 & 57.63 & 59.97 & 4.31 & 62.34 & 67.55 & 44.73  \\
        \cmidrule(l){2-9}
        & STA (Ours)     & 89.93 & 76.98 & 83.45 & 4.29 & 62.35 & 65.05 & 43.90 \\
        & wo/Amplitude     & 89.93 & 77.06 & 83.50 & 4.29 & 62.37 & 65.05 & 43.90 \\
        & wo/Frequency    & 87.26$\downarrow$ & 75.06$\downarrow$ & 81.16$\downarrow$ & 4.33 & 62.61 & 68.92 & 45.29 \\
        \midrule
        \midrule

        \multirow{4}{*}{\parbox{2cm}{\centering\textbf{Gemma-2-9b-it}}} 
        & Vanilla & 70.37 & 97.41 & 83.89 & 5.39 & 72.06  & 75.66  & 51.04 \\
        \cmidrule(l){2-9}
        & STA (Ours) & 95.78 & 99.33 & 97.56 & 5.43 & 70.27 & 71.65 & 49.12 \\
        & wo/Amplitude     & 95.70 & 99.33 & 97.52 & 5.43 & 70.29 & 71.49 & 49.07 \\
        &  wo/Frequency    & 90.89$\downarrow$ & 98.42$\downarrow$ & 94.65$\downarrow$ & 5.43 & 70.90 & 72.63 & 49.65 \\
        
        \bottomrule
        \end{tabular}
    }
    \caption{The ablation study of our proposal STA.
    The biggest drop of detoxification performance in each column is appended $\downarrow$.
    }
    \label{tab:ablation}
    }
\end{table*}

We remove the Amplitude component (wo/Amplitude) and the Frequency component (wo/Frequency) separately to analyze their individual contributions. 
As shown in Table \ref{tab:ablation}, removing Frequency leads to a greater drop in target capabilities compared to removing Amplitude. However, the effectiveness of Frequency relies on a larger amount of data; when data is limited, the Amplitude component becomes crucial for maintaining performance.

\section{Comparison to Paremter-tuning}
\label{appendix:tuning}
\definecolor{Mycolor1}{HTML}{BAD8F2}
\definecolor{Mycolor2}{HTML}{E8F2FB}
\definecolor{Mycolor3}{HTML}{FAE4E3}
\begin{table*}[ht]
    \centering
    \setlength{\tabcolsep}{3pt}
    {
    \resizebox{\linewidth}{!}{
        \begin{tabular}[c]{cc|ccc|cccc}
        \toprule
        \multirow{2}{*}{\textbf{Model}}
        & \multirow{2}{*}{\textbf{Method}}
        & \multicolumn{3}{c|}{\textbf{Detoxification Performance }}
        & \multicolumn{4}{c}{\textbf{General Performance}} \\
        \cmidrule(l){3-5}\cmidrule(l){6-9}
        & & {SafeEdit} & {RealToxicprompts} &{Avg} &{Fluency} &{MMLU} &{GSM8K} &{Avg} \\
        \midrule
        \multirow{7}{*}{\parbox{2cm}{\centering\textbf{Gemma-2-9b-pt}}}
        & Vanilla & 62.30 & 57.63 & 59.97 & 4.31 & 62.34 & 67.55 & 44.73  \\
        \cmidrule(l){2-9}
        & SFT & 68.44 & 58.47 &63.45  & 4.27 & \textbf{64.31} & \underline{69.07} &\underline{45.88}  \\
        & DPO & 81.48 & 58.05 &69.76  & 4.37 & \underline{64.19} & \textbf{69.83} &\textbf{46.13}  \\
        \cmidrule(l){2-9}
        & $\text{Prompt}_{hand}$ & 72.52 & 53.96 & 63.24 & 3.88 & 57.01 & 67.48 & 42.79 \\
        & $\text{Prompt}_{auto}$ & 64.15 & 57.63 & 60.89 & 4.19 & 60.09 & 68.61 & 44.30 \\
        \cmidrule(l){2-9}
        & CAA & \underline{85.78} & \underline{73.98} & \underline{79.88} & \textbf{4.38} & 61.35 & 68.54 & 44.76 \\
        & STA (Ours)     & \textbf{89.93} & \textbf{76.98} & \textbf{83.45} & 4.29 & 62.35 & 65.05 & 43.90 \\
        \midrule
        \midrule

        \multirow{7}{*}{\parbox{2cm}{\centering\textbf{Gemma-2-9b-it}}} 
        & Vanilla & 70.37 & 97.41 & 83.89 & 5.39 & 72.06  & 75.66  & 51.04 \\
        \cmidrule(l){2-9}
        & SFT & 91.41 & 97.83 &94.62  & 5.42 & \textbf{72.13} & \textbf{76.50} &\textbf{51.35}  \\
        & DPO & \textbf{98.52} & 98.42 &\textbf{98.47}  & 5.36 & 72.03 & 75.36 &50.92  \\
        \cmidrule(l){2-9}
        & $\text{Prompt}_{hand}$ & 78.74 & 98.42 & 88.58 & 5.41 & 71.07 & 74.83 & 50.44 \\
        & $\text{Prompt}_{auto}$ & 75.56 & 98.92 & 87.24 & \textbf{5.44} & 70.79 & 75.66 & 50.63 \\
        \cmidrule(l){2-9}
        & CAA & 91.48 & 98.75 & 95.12 & 5.42 & 70.77 & 75.21 & 50.47 \\
        & STA (Ours) & 95.78 & \textbf{99.33} & 97.56 & 5.43 & 70.27 & 71.65 & 49.12 \\
        \bottomrule
        \end{tabular}
    }
    \caption{The detoxification performance and its side effects on the general capabilities of parameter-tuning, prompting, and steering strategies. 
    The best results are marked in \textbf{bold} and the second-best results are marked with \underline{underline}. 
    }
    \label{tab:tuning}
    }
\end{table*}

We compare steering methods with parameter-tuning approaches (e.g., SFT and DPO). 
As shown in the Table \ref{tab:tuning}, steering strategies outperform SFT and DPO on Gemma-2-9B-pt.
However, on Gemma-2-9B-it, steering methods fall short compared to SFT and DPO. 
Note that steering is an inference-time intervention strategy and can be applied on top of models fine-tuned with SFT, DPO, or other parameter-tuning methods \cite{CAA}.
Additionally, as illustrated in Table \ref{tab:tuning}, steering strategies (CAA and our STA) consistently outperform prompting strategies.

\section{Personality}
\label{appendix:personality}

In the \textbf{personality domain}, we analyze LLM behavior on datasets \textit{ myopic reward} \cite{CAA,DBLP:conf/acl/PerezRLNCHPOKKJ23}.

\paragraph{STA can control personality behaviors of LLMs.}
We evaluate both steering and prompting strategies on the myopic reward personality trait. As shown in Table \ref{tab:personality}, the three steering strategies (CAA, $\text{SAE}_{AXBENCH}$, and STA), perform comparably across four metrics, all outperforming prompting-based methods.

\begin{table*}[ht]
\centering
\begin{tabular}{l|cccc}
\hline
\textbf{Method} & \textbf{Myopic} &\textbf{Fluency} &\textbf{MMLU} &\textbf{GSM8K}\\
\hline
        Vanilla &48 &4.07 &72.06  &75.66\\
        $\text{Prompt}_{auto}$ &64  &4.10 &71.59 &73.69 \\
        CAA &74  &4.07 &71.88 &76.95 \\
        $\text{SAE}_{AXBENCH}$ &74  &4.09 &71.77 &76.04 \\
        STA (ours) &74  &4.09 &71.74 &75.66 \\    
\hline
\end{tabular}
\caption{The performance on \textit{myopic reward} of  STA and baselines. 
}
\label{tab:personality}
\end{table*}

\section{Prompting and Steering}
\label{appendix:Prompting and Steering}

\subsection{Position of Prompt}
\label{appendix:position}
\begin{figure}[htbp]
    \centering
     \includegraphics[width=0.5\textwidth]{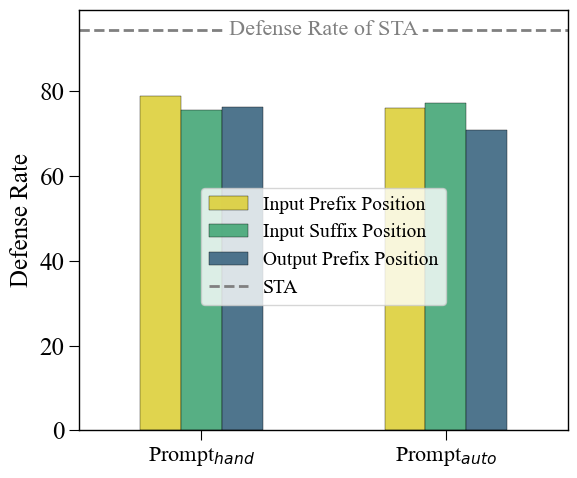}
    \caption{The detoxification performance and prompt at different positions.}
    \label{fig:position}
\end{figure}
We begin by selecting two competitive prompts: a manually designed prompt \(\text{Prompt}_{\text{hand}}\) \cite{self-reminders} and an automatically generated prompt \(\text{Prompt}_{\text{auto}}\) \cite{AxBench}.
To maximize their effectiveness, we concatenate these prompts at various positions, including the input prefix, input suffix, and output prefix. As illustrated in Fig \ref{fig:nums}, the performance of prompts varies significantly depending on their placement, with the optimal position differing between the two prompts. 
In Table \ref{tab:overall_performance}, we report results using the best-performing positions for each prompt. 
However, even with optimal placement, prompting fails to surpass the performance of STA, as demonstrated in Fig \ref{fig:position}.

\subsection{The performance of Prompting and Steering}
\label{appendix:The performance of Prompt}

\paragraph{The boundary of $\text{STA}_{prompt}$}

\begin{figure*}[!t]
    \centering
    \begin{subfigure}[b]{0.49\linewidth}
        \includegraphics[width=\textwidth]{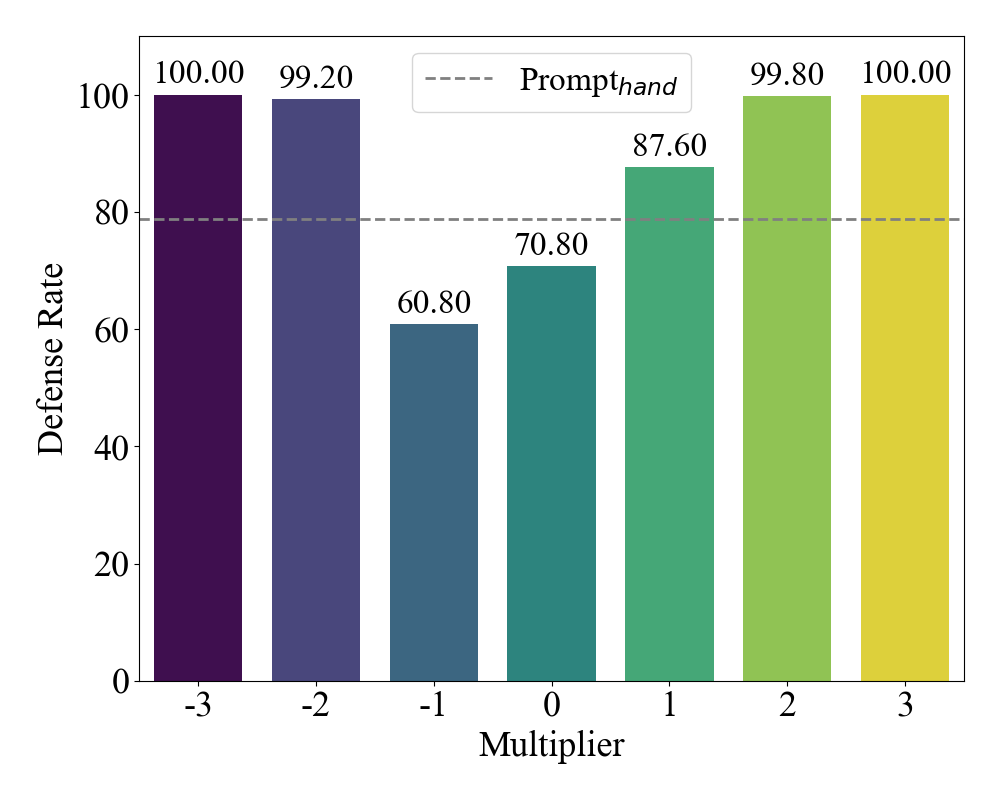}  
        \caption{The steering boundary of $\text{STA}_{prompt}$ unsing $\text{Prompt}_{hand}$.}
        \label{fig:hand_vector_boundary}
    \end{subfigure}
     \begin{subfigure}[b]{0.49\linewidth}
        \includegraphics[width=\textwidth]{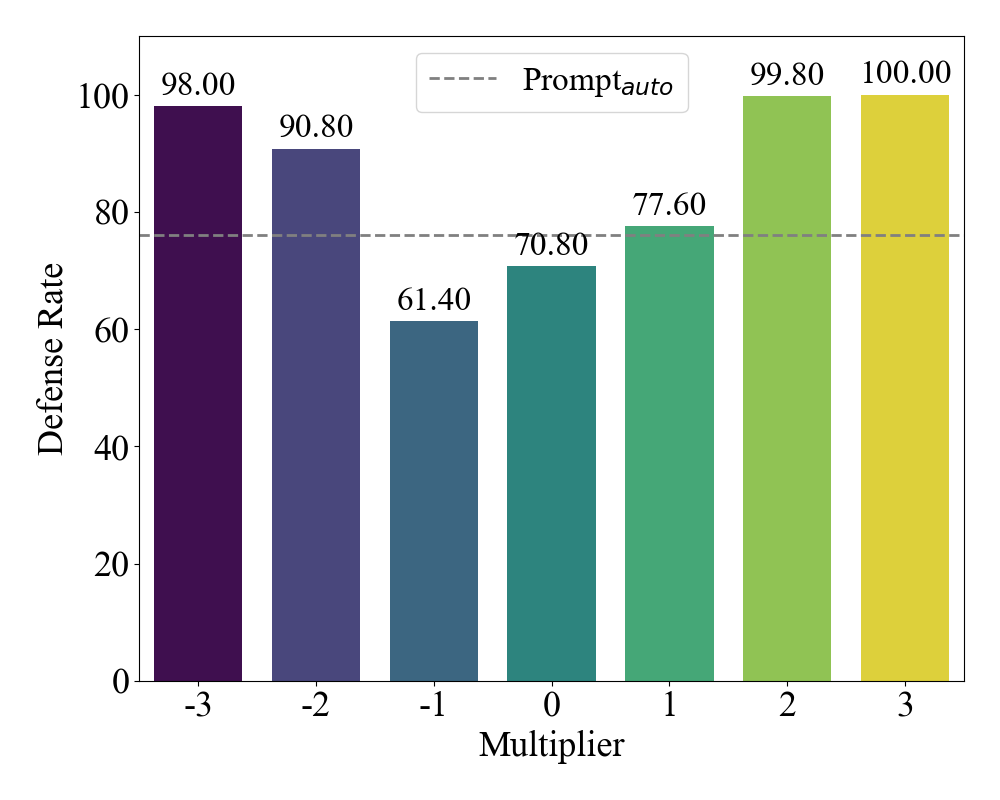}
        \caption{The steering boundary of $\text{STA}_{prompt}$ unsing $\text{Prompt}_{auto}$.}
        \label{fig:auto_vector_boundary}
    \end{subfigure}
    \caption{The controlling boundary on safety domian of prompting (few-shot demonstrations) and steering strategy.}
    \label{fig:prompt_to_vector_boundary}
\end{figure*}

We also analyzed the control capability of the steering vectors obtained by directly using the $\text{Prompt}_{hand}$ and $\text{Prompt}_{auto}$ transformations. 
Specifically, as shown in Fig \ref{fig:boundary}, the control range of $\text{STA}_{prompt}$ using $\text{Prompt}_{auto}$, with a multiplier ranging from -3 to +3, varies between -8.97\% and +29.63\%.

\section{Controlling the length of thought}
\label{appendix:Controlling the length of thought}
\subsection{Data}
\label{appendix:thought_data}
We construct an instance with both long thought and short thought answer:

\begin{center}
 \setlength{\fboxsep}{8pt} 
  \colorbox{gray!20}{\begin{minipage}{\dimexpr\linewidth-2\fboxsep} 
    \textbf{Question}: 1 + 1 =
  \end{minipage}}
\end{center}

\begin{center}
 \setlength{\fboxsep}{8pt} 
  \colorbox{gray!20}{\begin{minipage}{\dimexpr\linewidth-2\fboxsep} 
    \textbf{Answer} (\textbf{short thought}):
1 + 1 equals 2. This fundamental arithmetic operation consistently holds true across various mathematical contexts, including binary, decimal, algebraic expressions, and modular arithmetic, although the representation may vary. In standard arithmetic, the sum of two ones is always two.
  \end{minipage}}
\end{center}

\begin{center}
 \setlength{\fboxsep}{8pt} 
  \colorbox{gray!20}{\begin{minipage}{\dimexpr\linewidth-2\fboxsep} 
    \textbf{Answer} (\textbf{long thought}): Alright, so I'm trying to figure out what 1 + 1 equals. Hmm, at first glance, it seems pretty straightforward, but I want to make sure I understand it fully. Let me think about how addition works. When you add two numbers, you're combining their quantities. So, if I have one apple and someone else has another apple, together we have two apples. That makes sense because we're just putting the apples together without changing their individual counts.

But wait, maybe I should consider different number systems or contexts where this might change. For example, in binary, which is the base-2 system, 1 + 1 equals 10. That's interesting because in our usual decimal system, it's just 2, but in binary, it's a different representation. So, the way we add numbers can vary depending on the base we're using.

Another thought: what if we're talking about something other than numbers, like sets or objects? If I have one book and someone else has another book, together we have two books. It's the same concept, just adding the quantities. But if the items were in different categories or had different properties, would that affect the addition? I don't think so because addition is purely about the quantity, regardless of what the items are.

...

1 + 1 equals 2. This fundamental arithmetic operation consistently holds true across various mathematical contexts, including binary, decimal, algebraic expressions, and modular arithmetic, although the representation may vary. In standard arithmetic, the sum of two ones is always two.
  \end{minipage}}
\end{center}

\subsection{Setting}
\label{appendix:setting}
To assess the boundary of thought control via steering, we conducted experiments on 20 samples from the GSM8K dataset, running each sample 5 times and calculating the average length of the answer tokens. The temperature coefficient of DeepSeek-R1-Distill-Qwen-7B was set to 0.1, and due to resource constraints, we limited the max new tokens to 3000. However, we observed that when the steering coefficient was set to -2, the model tended to repeat solutions and, in fact, exceeded the 3000-token limit. More extensive experiments will be left for future work.

\end{document}